\documentclass[10pt,twocolumn,letterpaper]{article}

\usepackage[pagenumbers]{wacv} %

\usepackage{graphicx}
\usepackage{amsmath}
\usepackage{amssymb}
\usepackage{booktabs}
\usepackage{color}

\usepackage{booktabs}
\usepackage{multirow}
\usepackage{makecell}
\usepackage{adjustbox}
\usepackage{pifont} %
\usepackage{enumitem}
\setlist[itemize]{noitemsep,leftmargin=*,topsep=0em}
\setlist[enumerate]{noitemsep,leftmargin=*,topsep=0em}
\usepackage[table,xcdraw]{xcolor}
\usepackage{colortbl}
\usepackage{placeins}
\usepackage{times} 
\usepackage[absolute]{textpos}

\usepackage[pagebackref,breaklinks,colorlinks]{hyperref}

\usepackage[capitalize]{cleveref}
\crefname{section}{Sec.}{Secs.}
\Crefname{section}{Section}{Sections}
\Crefname{table}{Table}{Tables}
\crefname{table}{Tab.}{Tabs.}

\newcommand{\wpink}[1]{\textcolor{magenta}{#1}}
\newcommand{\fblue}[1]{\textcolor{black}{#1}}

\newcommand{\cmark}{\ding{51}}%
\newcommand{\xmark}{\ding{55}}%
\newcommand{\bnum}[1]{\bfseries #1}

\newcommand{\dname}{WeedsGalore}  %

\begin{document}

\title{\dname: A Multispectral and Multitemporal UAV-based Dataset for Crop and Weed Segmentation in Agricultural Maize Fields}

\begin{textblock}{13}(1.3, 0.5)
\begin{center}
  {\textcolor[HTML]{808080}{This paper has been accepted for publication at the IEEE/CVF Winter Conference on Applications of Computer Vision (WACV) 2025. $\copyright$ IEEE.}}
\end{center}
\end{textblock}

\author{Ekin Celikkan$^{1,2}$\and Timo Kunzmann$^{1}$\and Yertay Yeskaliyev$^{1}$\and Sibylle Itzerott$^{1}$\and Nadja Klein$^{3}$\and Martin Herold$^{1}$\and
$^1$GFZ German Research Centre for Geosciences\quad $^2$Humboldt-Universität zu Berlin \\ $^3$Scientific Computing Center, Karlsruhe Institute of Technology \\
\tt\small{\{ekin.celikkan, itzerott, herold\}@gfz.de}\quad\quad\tt\small{nadja.klein@kit.edu}
}
\maketitle

\begin{abstract}
Weeds are one of the major reasons for crop yield loss but current weeding practices fail to manage weeds in an efficient and targeted manner. Effective weed management is especially important for crops with high worldwide production such as maize, to maximize crop yield for meeting increasing global demands. Advances in near-sensing and computer vision enable the development of new tools for weed management. Specifically, state-of-the-art segmentation models, coupled with novel sensing technologies, can facilitate timely and accurate weeding and monitoring systems. However, learning-based approaches require annotated data and show a lack of generalization to aerial imaging for different crops. We present a novel dataset for semantic and instance segmentation of crops and weeds in agricultural maize fields. The multispectral UAV-based dataset contains images with RGB, red-edge, and near-infrared bands, a large number of plant instances, dense annotations for maize and \fblue{four} weed classes, and is multitemporal. We provide extensive baseline results for both tasks, including probabilistic methods to quantify prediction uncertainty, improve model calibration, and demonstrate the approach’s applicability to out-of-distribution data. The results show the effectiveness of the two additional bands compared to RGB only, and better performance in our target domain than models trained on existing datasets. We hope our dataset advances research on methods and operational systems for fine-grained weed identification, enhancing the robustness and applicability of UAV-based weed management. The dataset and code are available at \href{https://github.com/GFZ/weedsgalore}{https://github.com/GFZ/weedsgalore.}

\end{abstract}

\section{Introduction}
\label{sec:intro}

\begin{figure}
    \centering
    {\includegraphics[trim={3.5cm 1.5cm 11cm 3cm},clip,width=1\linewidth]{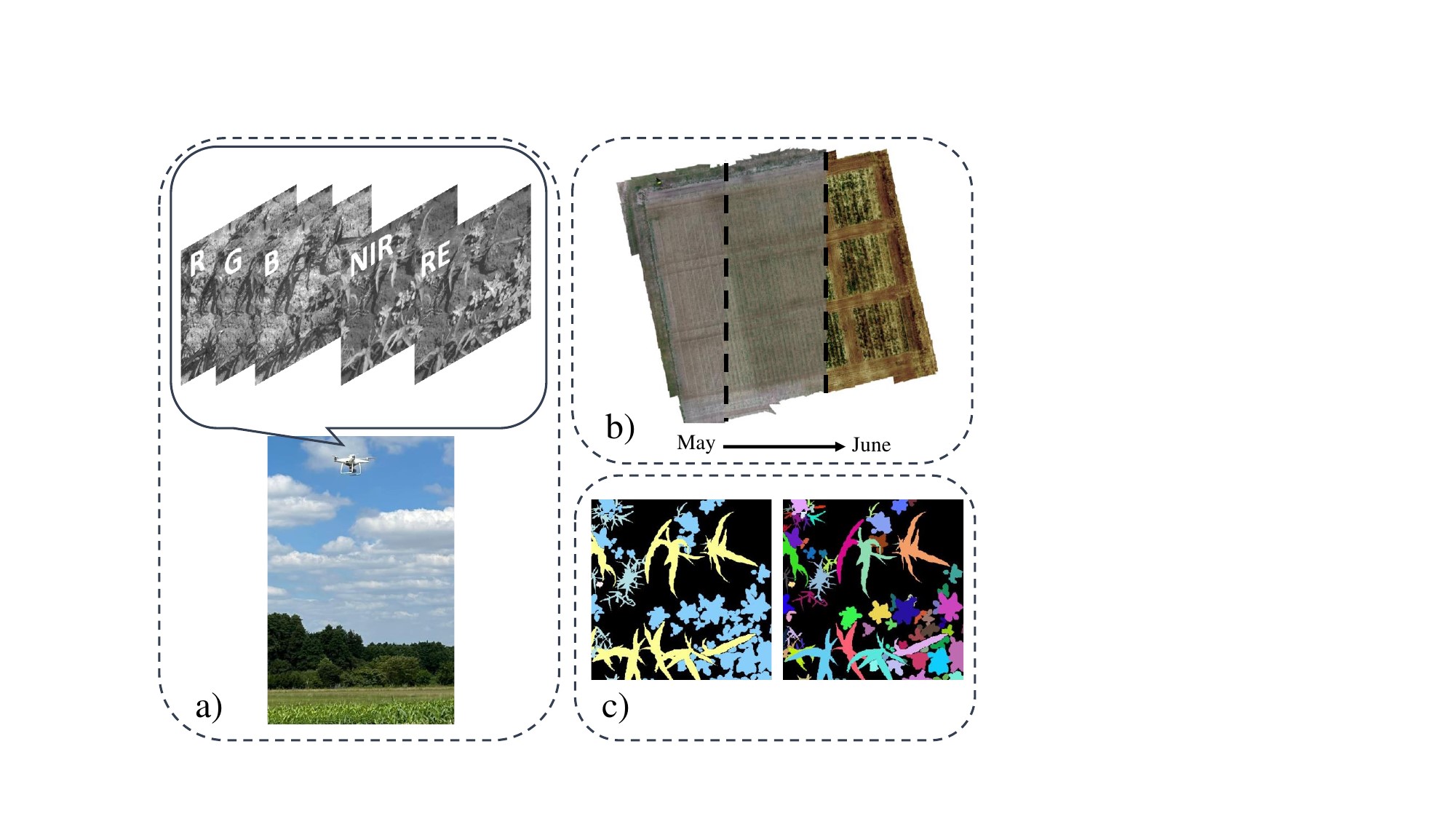}}
    
    \caption {\textbf{\textit{WeedsGalore} dataset.} We present a novel reference dataset for UAV-based weed monitoring for maize fields. The dataset a) contains images with 5-bands (RGB, near-infrared, red-edge) and b) is recorded at different growth stages, fully reflecting real-world agricultural practices. c) We provide detailed pixel-level annotations for semantic (multiple weed classes) and instance segmentation.}
    \label{fig:eyecatcher}
\end{figure}

Weeds are a major reason for crop yield loss as they compete with desired crops for resources like nutrients, water, and space \cite{agronomy10040466}.
The global demand for food is constantly rising, while climate change and economic considerations pose challenges to agricultural production \cite{Westwood_Charudattan_2018,agronomy12010118}.
Hence, any factor threatening the yield should be managed and minimized, one such factor being weeds.

There are two common approaches to weeding: Chemical (i.e.~spraying herbicides) and mechanical (i.e.~physical removal) weeding. Both treatments have negative environmental impacts and financial costs. These are further amplified by excessive usage of chemicals, leading to herbicide-resistant weed populations \cite{https://doi.org/10.1002/ps.4821}. In response to those challenges, site-specific weed management (SSWM) aims to target weeding to specific areas. To be able to realize SSWM effectively and at scale, as is the case in real-world agricultural fields, automated systems and data with high temporal and spatial precision are crucial.

Unmanned aerial vehicle (UAV) based systems are one of the most promising directions in agriculture. They do not suffer from the drawbacks associated with ground-based vehicles that have limited viewpoints and the inability to navigate complex terrains. This makes UAVs ideal candidates for weed monitoring.
However, approaches relying on existing datasets fail to transfer to UAV systems for unseen crop types. Maize fields, having the highest worldwide production in the cereal category \cite{FAO2023}, are such a domain, for which such automated systems are crucial for future agricultural practices. There is very limited publicly available data for weed segmentation in general, especially when it comes to UAVs and more advanced sensors beyond RGB, which contain extra information useful for plant imaging \cite{milioto2018real, WOLFF2023110140}.
We introduce \textit{\dname}, a dataset for pixel-level identification of maize and weeds.
To the best of our knowledge, it is the first UAV-based dataset providing semantic and instance labels for crops and weeds in maize fields.
We use a non-invasive UAV for data acquisition and provide labels for \fblue{four} weed classes, where the multitemporality is incorporated in terms of diversity in growth stages.

Segmentation models are useful beyond SSWM and there is an increasing interest in estimating the weed cover for plant science research,
where computer vision-based weed localization can help in testing and comparing treatments (e.g. herbicides).
A typical approach is to conduct different experiments regarding herbicide applications and calculate the weed control efficiency of the tested herbicide \cite{Patel_Jha_Verma_Porwal_Toppo_Raghuwanshi_2023, plants13020211}.
Traditionally, the weed cover during these experiments has been measured manually, which is naturally very cumbersome and error-prone \cite{sairam2023effect, Wang_2023_ICCV}.
Hence, models that can accurately segment weeds are vital to support and accelerate agricultural research.
To tackle the current lack of publicly available datasets for fine-grained weed segmentation in maize fields, our dataset \dname~ provides images with five bands (red, green, blue (RGB), red-edge (RE), near-infrared (NIR)) and contains four different weed classes, enabling monitoring of different weed populations.

Model selection plays an integral role in the real-world deployment of automated weed segmentation methods since real-world data is typically confronted with unseen objects and noisy measurements.
As a result, models that are well-calibrated and quantify prediction uncertainty realistically are needed, to correctly detect overconfident predictions and areas of high uncertainties.
Both properties can significantly contribute to risk management and generalizability in real-world agricultural fields.
To meet the need for reliable methods, we propose to deploy a Bayesian model with uncertainty quantification on our dataset.

We evaluate \textit{\dname} for the tasks of semantic and instance segmentation. The results show that our dataset uniquely enables the application of aerial monitoring in maize fields, and through the use of probabilistic methods the model performance and calibration can be significantly improved, especially on out-of-distribution (OOD) data.

In summary, our contributions are:
\begin{enumerate}
    \item We present a novel reference dataset for crop and weed segmentation in maize fields, with dense annotations for multiple weed species covering multiple phenological stages. To the best of our knowledge, it is the first publicly available multispectral UAV dataset for weed monitoring in maize fields, offering two extra bands compared to conventional RGB datasets and with unprecedented weed density.

    \item We provide extensive quantitative evaluation for semantic and instance segmentation with several state-of-the-art methods, including probabilistic methods for improved model calibration and uncertainty quantification.

    \item We furthermore demonstrate the usefulness of our dataset by testing our method under real-world conditions and show that models trained on \dname~can be successfully deployed in unseen maize fields, including large-scale orthomosaics.

\end{enumerate}

\section{Related Work}
\label{sec:related_work}
\definecolor{lightyellow}{RGB}{255, 255, 204}
\definecolor{lightred}{RGB}{255, 204, 204}
\definecolor{lightgreen}{RGB}{204, 255, 204}

\begin{table*}[t!]
    \centering
    \adjustbox{max width=0.95\textwidth}{%
    \setlength{\tabcolsep}{4pt}
    \begin{tabular}{cccccccccccccr}
        \toprule
        \multirow{2}{*}[-4pt]{Dataset}                         & %
        \multirow{2}{*}[-4pt]{Year}                            & %
        \multirow{2}{*}[-4pt]{Crop}                            & %
        \multirow{2}{*}[-4pt]{Platform}                        & %
        \multirow{2}{*}[-4pt]{GSD [$\frac{px}{mm}$]}           & %
        \multirow{2}{*}[-4pt]{Modality}                        & %
        \multicolumn{2}{c}{Annotations}                        & %
        \multirow{2}{*}[-4pt]{\#weed classes}                  & %
        \multicolumn{2}{c}{\#Instances~/~Image}                & %
        \multirow{2}{*}[-4pt]{Multitemp.}                      & %
        \multirow{2}{*}[-4pt]{\fblue{Resolution}}              \\ %
        \cmidrule(lr){7-8}
        \cmidrule(lr){10-11}
                                                                & %
                                                                & %
                                                                & %
                                                                & %
                                                                & %
                                                                & %
        Semantic                                                & %
        Instance                                                & %
                                                                & %
        Crop                                                    & %
        Weed                                                    & %
                                                                & %
                                                                \\ %
        \midrule
        Carrot-Weed \cite{haug15}                               & %
        2015                                                    & %
        Carrot                                                  & %
        UGV                                                     & %
        8.95                                                    & %
        MSI                                                     & %
        \cmark                                                  & %
        \xmark                                                  & %
        1                                                       & %
        {--}                                                    & %
        {--}                                                    & %
        \xmark                                                  & %
        $1296 \times 966$                                       \\ %
        WeedMap \cite{rs10091423}                               & %
        2018                                                    & %
        Sugar beet                                              & %
        UAV                                                     & %
        $>$8.3                                                  & %
        MSI                                                     & %
        \cmark                                                  & %
        \xmark                                                  & %
        1                                                       & %
        {--}                                                    & %
        {--}                                                    & %
        \cmark                                                  & %
        {--}                                                    \\ %
        CoFly-WeedDB \cite{KRESTENITIS2022108575}               & %
        2022                                                    & %
        Cotton                                                  & %
        UAV                                                     & %
        {--}                                                    & %
        RGB                                                     & %
        \cmark                                               & %
        \xmark                                               & %
        3                                                       & %
        {--}                                                    & %
        {--}                                                    & %
        \xmark                                                  & %
        $1280 \times 720$                                       \\ %
        Sorghum-Weed \cite{GENZE2022107388}                     & %
        2022                                                    & %
        Sorghum                                                 & %
        UAV                                                     & %
        1.0                                                     & %
        RGB                                                     & %
        \cmark                                                  & %
        \xmark                                                  & %
        1                                                       & %
        {--}                                                    & %
        {--}                                                    & %
        \xmark                                                  & %
        {--}                                                    \\ %
        WE3DS \cite{s23052713}                                  & %
        2023                                                    & %
        7*                                                      & %
        UGV                                                     & %
        0.4                                                     & %
        RGB-D                                                   & %
        \cmark                                                  & %
        \xmark                                                  & %
        10*                                                     & %
        {--}                                                    & %
        {--}                                                    & %
        \cmark                                                  & %
        $1600 \times 1144$                                      \\ %
        PhenoBench \cite{weyler2023phenobench}                  & %
        2023                                                    & %
        Sugar beet                                              & %
        UAV                                                     & %
        1.0                                                     & %
        RGB                                                     & %
        \cmark                                                  & %
        \cmark                                                  & %
        1                                                       & %
        8.55                                                    & %
        5.70                                                    & %
        \cmark                                                  & %
        $1024 \times 1024$                                      \\ %
        CropAndWeed (Fine24) \cite{Steininger_2023_WACV}        & %
        2023                                                    & %
        8*                                                      & %
        hand-held                                               & %
        {--}                                                       & %
        RGB                                                     & %
        \cmark                                               & %
        \xmark                                               & %
        16*                                                     & %
        {--}                                                    & %
        {--}                                                    & %
        \cmark                                                  & %
        $1920 \times 1088$                                      \\ %
        \midrule
        \textbf{\dname~(Ours)}                                  & %
        2024                                                    & %
        Maize                                                   & %
        UAV                                                     & %
        2.5                                                     & %
        MSI                                                     & %
        \cmark                                                  & %
        \cmark                                                  & %
        4                                                       & %
        13.91                                                   & %
        64.30                                                   & %
        \cmark                                                  & %
        $600 \times 600$                                        \\ %
        \bottomrule
    \end{tabular}
    }
    \caption{ {\bf Comparison of publicly available real-world agricultural datasets that provide pixel-level annotations for weed and crop segmentation.} GSD refers to ground sampling distance.
    *Multi-class datasets (i.e.~multiple weed and crop classes).
    }
     \label{tab:dataset_comparison}
\end{table*}

\subsection{Weed Segmentation for Precision Agriculture and Plant Science Research}

As in many domains, computer vision techniques have found interest in agricultural applications and research. One such task is weed monitoring, as there is growing interest in developing automated weeding systems that effectively localize and remove weeds from agricultural croplands \cite{Champ20aps, https://doi.org/10.1002/rob.21938, DENG2024102546}.
Existing approaches can be analyzed based on their choices for system components, such as imaging platform, sensor/image modality, target application (i.e.~crop type), task formulation, and model selection.  

\noindent \textbf{Imaging platform.}
The majority of existing systems rely on unmanned ground vehicles (UGVs), although they have many limitations,
for example, the inherent risk of damaging plants, the need for mechanical adjustment to different environments, and the deployment on crop fields with minimal inter- and intra-row distances (e.g. wheat, barley, etc. \cite{DEVITA201769, BOSTROM2012144}).
A promising alternative is using images acquired by UAVs (i.e., drones), and localizing and classifying plants of interest.
Our system and data are UAV-based, as UAVs stand out due to their agile and non-invasive nature, and commercial availability \cite{9705188, 8239328}.

\noindent \textbf{Imaging sensor.} The overwhelming majority of existing work on weed segmentation relies only on RGB data \cite{GENZE2022107388, weyler2023phenobench, Steininger_2023_WACV, app10207132}, which is surprising as more bands hold useful information for weed segmentation \cite{9081915, milioto2018real}. We utilize multispectral imagery (MSI), which contains two extra bands compared to regular RGB sensors. 

\noindent \textbf{Target crop.} \cite{weyler2023phenobench, Sa2018}, \cite{KRESTENITIS2022108575} and \cite{Genze2023} use UAV-based images for weed segmentation, but they are for sugarbeet, cotton and sorghum fields respectively. \cite{agronomy13071846, Steininger_2023_WACV} segment weeds in maize fields, but from images acquired by hand-held cameras. In this work, we present a system specifically tailored for UAV-based weed segmentation in maize fields.

\noindent \textbf{Task formulation.} The task is often formulated as a semantic or instance segmentation problem \cite{Ahmadi22iros, Champ20aps, milioto2018real, rs10091423}.
Given an input image, the model outputs a prediction (or a full weed map if the whole field is given as input \cite{rs10091423}) indicating per-pixel semantic classes or instance masks.
While the former is more fitting for targeted spraying (as it concerns larger patches, and the species information is useful to decide on the herbicide), and the latter is well-suited for mechanical weeding (where individual plants are removed); it is possibly the most effective when both are used in combination. We provide a reference dataset with both semantic and instance annotations, which can be used separately or in combination based on the target task.

\noindent \textbf{Model selection.}
Early work on weed segmentation heavily relies on classical machine learning methods with multi-stage pipelines and manual feature extraction~\cite{7989347, GUERRERO201211149, 6835733}.
Essentially, all recent approaches utilize state-of-the-art deep-learning models for weed and crop segmentation.
Milioto \etal \cite{milioto2018real} propose an encoder-decoder architecture for pixel-wise segmentation of sugarbeet crops and weeds.
\cite{agronomy13071846} use an improved Swin-Unet, and \cite{PICON2022106719} the PSPNet \cite{Zhao_2017_CVPR} (Pyramid Scene Parsing Network) for semantic segmentation of maize and weed. 
\cite{Wang_2023_ICCV} and \cite{rs10091423} deploy several CNN-based architectures on large-scale orthomosaics for wheat and sugar beet respectively.
Weyler \etal \cite{weyler2023phenobench} benchmark various methods like Mask R-CNN \cite{he2017mask} and Mask2Former \cite{cheng2021mask2former} for panoptic segmentation of sugarbeet and weed.
Recently there has been efforts for domain generalized methods for crop and weed segmentation to perform well under different field conditions, however, those methods focus on generalizing to different fields of the same crop and same acquisition mode \cite{Weyler2023TowardsDG, GAO2024122980}. Our dataset can be used complementary with other existing datasets for cross-domain validation or improved generalization.

\subsection{Weed Segmentation Datasets}
One of the driving forces behind progress in computer vision has been the availability of annotated data.
Unfortunately,
the amount and diversity of agricultural datasets are limited compared to more general datasets~\cite{Lin14eccv,Cordts16cvpr}
due to several factors, like the complexity of scenes (e.g.~complex illumination conditions, occlusion, overlap of plant organs) and the need for expert knowledge for annotations.

A comparison of existing datasets is provided in \cref{tab:dataset_comparison}. \cite{haug15} is an early carrot dataset that suffers from low ground resolution.
CoFly-WeedDB \cite{KRESTENITIS2022108575} contains RGB images captured on a single day at a cotton field in Greece.
It provides semantic masks only for the weeds, but not the crops.
PhenoBench \cite{weyler2023phenobench} is a UAV dataset, with pixel-level annotations for semantic and instance segmentation of sugarbeets and uni-class weeds.
The average number of plants per image is low, indicating a lack of challenging scenes and weed infestations.
In contrast, our \dname~contains around 64 weeds per image, compared to 5 of PhenoBench. WeedMap \cite{rs10091423} provides large-scale multispectral orthomosaics with semantic annotations for a uni-weed class and sugarbeet crops, but does not provide raw drone images.
Furthermore, the ground resolution is low, limiting the annotation detail and the applicability potential of the dataset.

While less common, there have been recent datasets that differentiate multiple weed classess. WE3DS \cite{s23052713} is an RGB-D dataset, collected with  UGV.
It contains multiple crop and weed species. However, the field (near Vienna, Austria) is a controlled research farm, and high-weed density areas are left out of the dataset, which does not fully reflect the real-life farming scenario.
CropAndWeed Dataset \cite{Steininger_2023_WACV} is a collection of RGB images from over a hundred locations in Austria.
The images are collected manually with a hand-held camera, which is not a suitable acquisition mode for automated systems in large-scale fields. \cref{tab:dataset_comparison} highlights the need for a UAV-based crop-weed segmentation dataset targeting maize fields, with challenging real-world scenes, which is addressed by our dataset.

\subsection{Probabilistic Deep Learning and Uncertainty Quantification for Semantic Segmentation}
\label{sec:probabilistic_dl}
The majority of the existing state-of-the-art models are deterministic. As a result, during test time, the model outputs a point estimate only. However, accurate uncertainty quantification (UQ) via e.g., prediction intervals has recently shown to be useful if not necessary to deploy vision algorithms in real-world automated systems \cite{9506719, Sagar_2022_WACV, HERNANDEZ2020106597, Fang_2020}.

One approach to quantify model confidence is to rely on softmax scores.
However, this is not always a reliable metric as uncertain models can output high softmax scores \cite{pmlr-v48-gal16, pmlr-v162-postels22a}.
An alternative is a Bayesian approach, assuming probability distributions over the network weights.
Commonly, variational inference (VI) with Monte Carlo (MC) dropout \cite{pmlr-v48-gal16} and Bernoulli distribution as the variational density is used to provide approximations to the true posterior distribution.
As a non-Bayesian alternative, deep ensembles \cite{NIPS2017_9ef2ed4b} have gained popularity, where the same model is trained multiple times with different initializations, and the different samples during inference are used to calculate predictive uncertainty. However, existing work for precision agriculture or autonomous weeding systems typically does not include uncertainty scores in their systems \cite{Champ20aps}.
Exceptions include \cite{vanmarrewijk2024active}, who consider epistemic uncertainty within the acquisition function in an active learning setting; and \cite{Celikkan_2023_ICCV}, who report predictive uncertainty scores for semantic segmentation of crops and weeds on the Sugarbeets2016 \cite{chebrolu2017ijrr} dataset.
In this paper, we deploy a probabilistic method, which not only improves segmentation performance but also reports reliable uncertainty scores.
We believe this information is vital to deploy perception algorithms in real-world robotic weeding systems. 

\section{The WeedsGalore Dataset}
\label{sec:method}
We provide a high-resolution UAV dataset for weed segmentation in maize fields.
In comparison to existing datasets, we provide extra input bands through multispectral sensing, dense semantic and instance annotations, and offer by far the largest number of weed instances per image.

\subsection{Data Collection}
\noindent\textbf{Field Description.}
The field is located in Marquardt, Potsdam, Germany (52°27'50.6"N 12°57'27.5"E) and covers an area of approximately 1840m$^2$.
To represent realistic agricultural scenarios, the farmers carried out the usual practices without experimental interference, and weeds occurred naturally.
The maize (\textit{Zea mays L.}) was planted across the entire field, on May 9, 2023.
The distance between rows was 70 cm with a 20 cm intra-row distance between crops.
Due to dry conditions, irrigation was applied on June 6.
Herbicide treatment was done on June 12.
Weeds were abundant in quantity as well as species diversity, the most dominant being common amaranth (\textit{Amaranthus retroflexus}).
Another species with a strong presence was a grass-like weed barnyard grass (\textit{Echinochloa crus-gall}), with quickweed (\textit{Galinsoga parviflora}) being the least common.
Apart from those three types of weed, there was a variety of other species in substantially fewer quantities.

\noindent\textbf{Data Acquisition.}
The dataset contains images from different dates, hence at different growth stages, plant cover, and weather conditions.
We carried out four data recording campaigns at dates May 25, May 30, June 6, and June 15.
The first campaign was when the first leaves of maize crops appeared (i.e., V1 stage \cite{yue2020prediction}) where there were almost no weeds present, while the last campaign had almost full plant cover (both from weeds and crops).

The images were taken with the DJI Phantom P4 Multispectral, a UAV equipped with five monochromic sensors for multispectral imaging (Blue: 450±16 nm, Green: 560±16 nm, Red: 650±16 nm, Red-Edge: 730±16 nm, Near-Infrared: 840±26 nm) with effective pixel of 2.08 MP as well as an RGB sensor \cite{djiMultispectralSpecifications}.
The flights were planned with DJI GS Pro \cite{djigspro}.
The UAV flew with an overlap of 70\% side and 60\% front overlap, and images were captured in Hover\&Capture mode to ensure the highest possible resolution.
The flight height was 5 meters to be able to distinguish individual plant instances, which resulted in a GSD (ground sampling distance) of 2.5mm.
As a result, approximately 1150 images were taken by the campaign.

Instead of creating the orthomosaic and annotating the orthophotos, our dataset contains raw images captured by the drone.
This has multiple reasons.
Firstly, similar to what was reported by previous work \cite{weyler2023phenobench}, we have observed the alignment causes artifacts, and this level of artifacts would lead to errors in our fine-grained segmentation masks.
Secondly, a model that is trained on raw images can be easily deployed in the future by spraying robot drones that work in real-time on the captured images, rather than relying on a multi-step orthorectification process.
\fblue{Hence, we followed these steps: First, the five single-band images were aligned (which corresponds to a translation of few pixels \cite{dji2020p4multispectral}). Second, the central 600x600 pixels (a square format that is flexible for processing, e.g.~data augmentation) were cropped, which were finally annotated. There is no overlap between annotated images. They are sampled from 48 separate locations as shown in \cref{fig:dataset_orthom}}. Yet, to have a clear overview of the field, and to structure the annotation process, we created the orthomosaic for each campaign using the software Agisoft Metashape \cite{agisoftAgisoftMetashape}.

\subsection{Data Annotation}
\begin{figure}
    \centering
    {\includegraphics[trim={4.8cm 2.6cm 9.3cm 3cm},clip,width=\columnwidth]{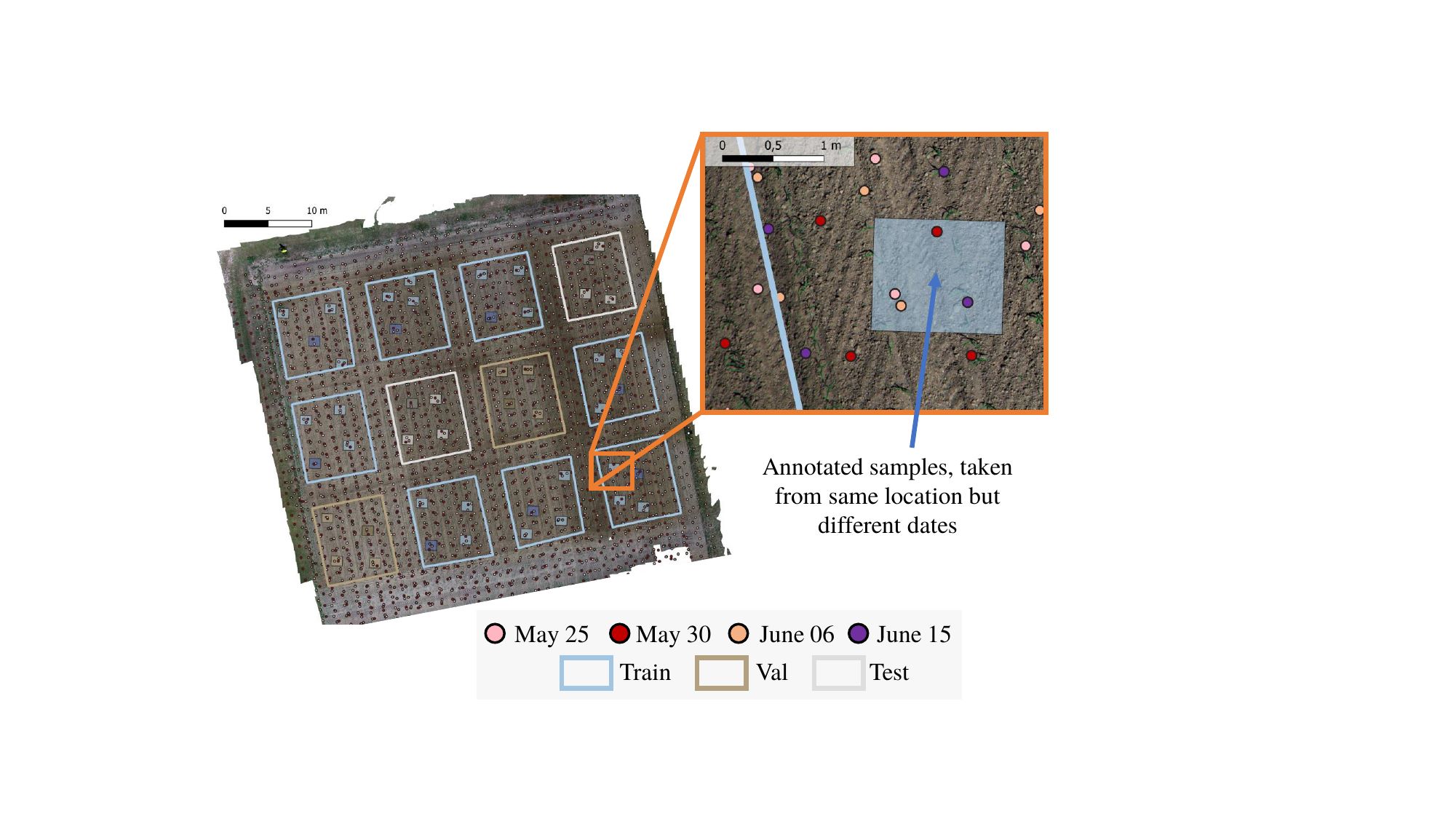}}
    \caption{\textbf{Ortohomosaic of the full field and georeferenced locations of annotated images.} Points show captured image locations, while their colors encode acquisition dates. The dataset splits are spatially separated by patches into \textcolor[HTML]{6a8caf}{train}, \textcolor[HTML]{e89c71}{validation}, and \textcolor[HTML]{a8a8a8}{test}. Smaller polygons are drawn around the annotated samples: The \textit{multitemporal} dataset contains samples from the same locations from all dates. Best viewed on screen and zoomed in.}
    \label{fig:dataset_orthom}
\end{figure}

To ensure the representation of different locations within the field, we divided the whole field into 12 patches with equal areas, and randomly sampled 4 geo-referenced point locations from each patch, resulting in 48 locations (see \cref{fig:dataset_orthom}).
36 of them were annotated for the first three recording dates (which are before the herbicide treatment, hence more relevant for the weed identification task as the growth stages of weeds on June 15 is too late for herbicide application). The remaining 12 were annotated for all four dates (marked with darker polygons in \cref{fig:dataset_orthom}), to also provide data for late-stage weeds, which can be used for either mechanical weeding or weed cover analysis. \fblue{The total annotated ground area corresponds to approximately 128 m$^2$ (per date).} Example scenes from four different dates are shown in \cref{fig:sample_scenes}.

\global\long\def\figWidth{0.2\linewidth}

\definecolor{crop}{HTML}{caca78}
\definecolor{weed1}{HTML}{87CEFA}
\definecolor{weed2}{HTML}{B0E0E6}
\definecolor{weed3}{HTML}{ccc0c4}
\definecolor{weed_other}{HTML}{CCCCFF}

\begin{figure}[ht!]
	\centering
    {\scriptsize
    \setlength{\tabcolsep}{2pt}
	\begin{tabular}{
	>{\centering\arraybackslash}m{0.1cm} 
	>{\centering\arraybackslash}m{\figWidth} 
	>{\centering\arraybackslash}m{\figWidth} 
	>{\centering\arraybackslash}m{\figWidth} 
	>{\centering\arraybackslash}m{\figWidth}
}

	&\includegraphics[clip,trim={{0cm 0cm 0cm 0cm}},width=\linewidth]{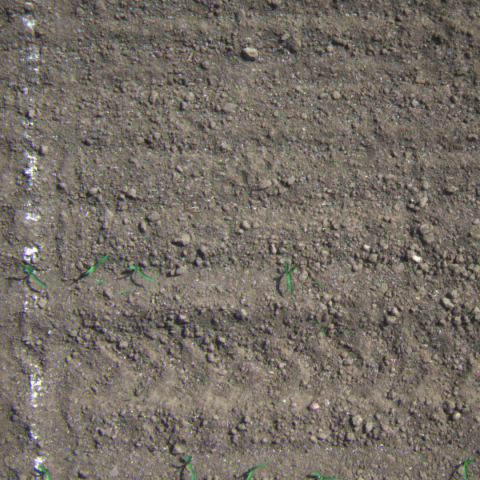}
        &\includegraphics[clip,trim={{0cm 0cm 0cm 0cm}},width=\linewidth]{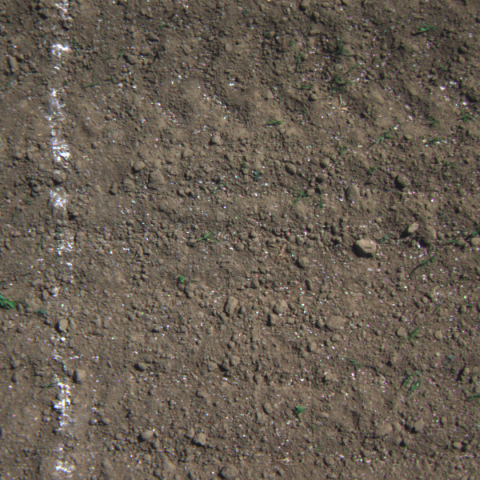}
        &{\includegraphics[clip,trim={{0cm 0cm 0cm 0cm}},width=\linewidth]{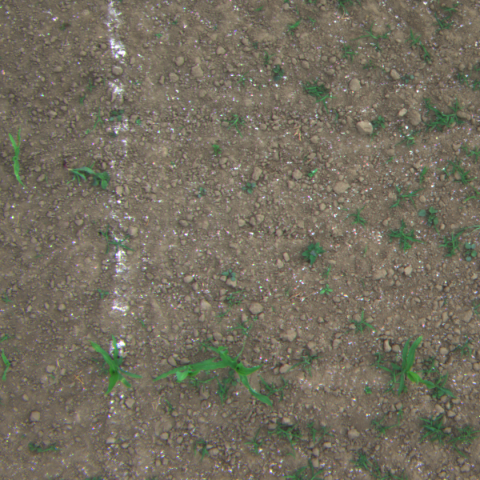}}
         &{\includegraphics[clip,trim={{0cm 0cm 0cm 0cm}},width=\linewidth]{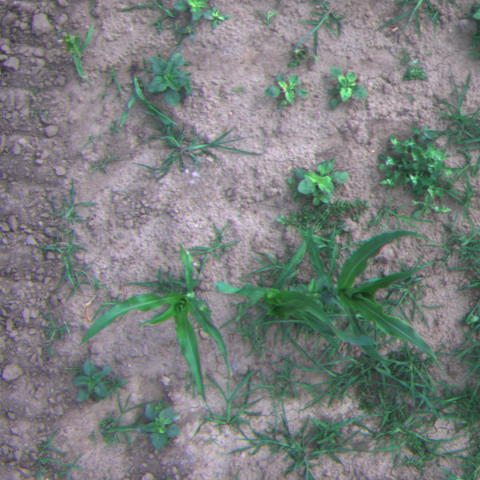}}
        \\

	&{\includegraphics[clip,trim={{0cm 0cm 0cm 0cm}},width=\linewidth]{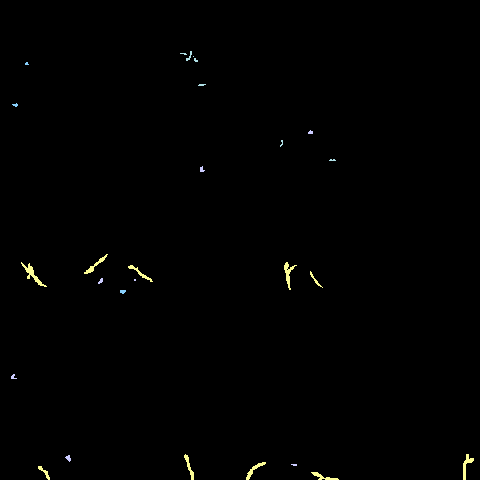}}
        &{\includegraphics[clip,trim={{0cm 0cm 0cm 0cm}},width=\linewidth]{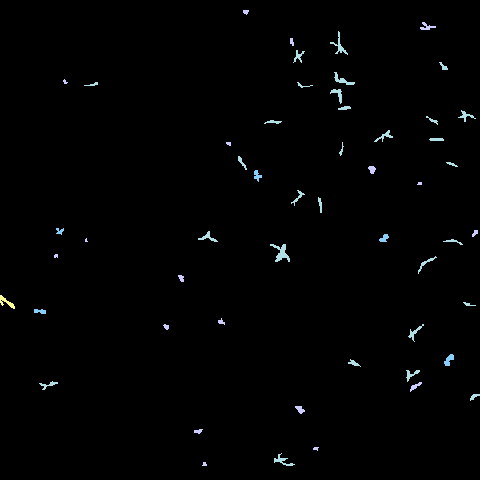}}
        &{\includegraphics[clip,trim={{0cm 0cm 0cm 0cm}},width=\linewidth]{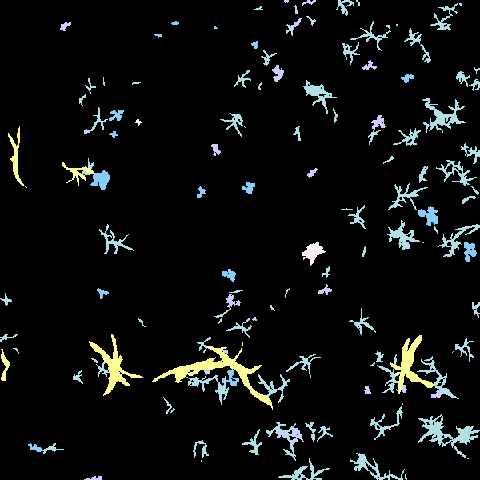}}
        &{\includegraphics[clip,trim={{0cm 0cm 0cm 0cm}},width=\linewidth]{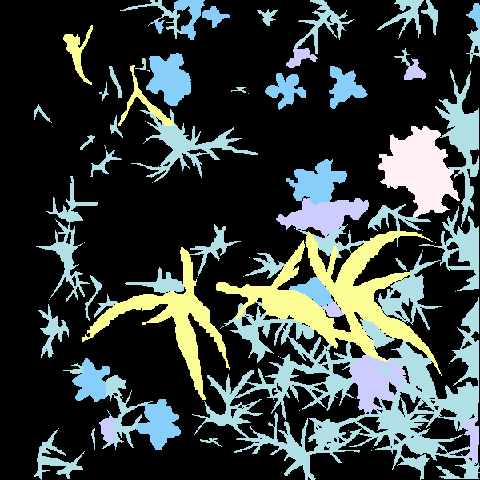}}
        \\

        & \raisebox{0pt}{May 25}
        & \raisebox{0pt}{May 30}
        & \raisebox{0pt}{June 6}
        & \raisebox{-4pt}{June 15}

	\end{tabular}
	}
  \caption{\textbf{Change in plant cover over the acquisition timeline.} Examples from four different dates and semantic masks (\textbf{\textcolor{crop}{maize}, \textcolor{weed1}{amaranth}, \textcolor{weed2}{barnyard grass}, \textcolor{weed3}{quickweed}, \textcolor{weed_other}{weed other})}. Best viewed on a colored screen, and zoomed in.}
  \label{fig:sample_scenes}
\end{figure}

The images were annotated using the web-based tool Encord \cite{encordCompleteData}. We used the AI-assisted tool with SAM \cite{kirillov2023segany} point prompt option, where the model made the initial mask prediction based on the click, which the annotators then refined. Hence, a total of 156 images were annotated by two annotators and mutually reviewed by three experts to ensure a high quality of the dataset, where the object boundary of each plant instance was marked with a polygon (i.e. instance mask) which was also assigned to one of the semantic classes from \textit{maize}, \textit{amaranth}, \textit{barnyard grass}, \textit{quickweed}, or \textit{weed other} (in the case of rarely grown species or when it was not possible to distinguish the class due to similar phenological appearance, especially during early growth stages). 

\subsection{Dataset Statistics}
We spatially split the data as 70:15:15 (i.e.~8/2/2 patches), such that images from the same location do not appear in more than one split, even if the data was recorded at a different date.
\cref{fig:dataset_orthom} shows the splits and corresponding locations in the orthomosaic, and the resulting dataset statistics can be seen from \cref{tab:dataset_stats}.
In addition to the annotated single images, we include 4 large orthomosaics in our dataset, providing a large amount of unlabeled data, which can be utilized for further research (\eg unsupervised learning, active learning, plant analysis, etc.) \cite{vanmarrewijk2024active, 10175653, DOSSANTOSFERREIRA2019104963}.

\begin{table}[t!]
    \centering
    \adjustbox{max width=\columnwidth}
    {\small{
    \begin{tabular}{@{}lcc|cccc|c}
      \toprule
      Split & \#crops & \#weed & \#amaranth & \#grass & \#quickweed & \#weed\_other & \fblue{\#images} \\
      \midrule
       Train & 1,461 & 6,512 & 2,618 & 1,813 & 161 & 1,920 & 104 \\
       Validation & 451 & 1,808 & 857 & 258 & 57 & 636 & 26 \\
       Test & 257 & 1,711 & 634 & 466 & 31 & 580 & 26 \\
       \midrule
      Total & 2,169 & 10,031 & 4,109 & 2,537 & 249 & 3,136 & 156 \\
      \bottomrule
    \end{tabular}
    }}
    \caption{\textbf{Dataset statistics.} The number of plant instances for each class within the defined splits are reported.}
    \label{tab:dataset_stats}
  \end{table}

\section{Experiments}
\label{sec:exp}
We provide extensive evaluation and provide baselines for our new dataset.
\cref{subsubsec:semantic} and \cref{subsubsec:instance} contain results for semantic and instance segmentation for different input data modalities.
\cref{subsec:exp:uq} presents UQ results on our dataset.
\cref{subsec:generalization} presents results on the generalization capabilities of models trained on our dataset, compared to existing ones.

\subsection{Benchmarks}
\label{subsec:benchmarks}
\subsubsection{Semantic Segmentation}
\label{subsubsec:semantic}
\noindent\textbf{Implementation Details and Evaluation Metrics.}
The goal of semantic segmentation is to assign pixel-level class labels. We evaluate two settings: 3-class (i.e., background, crop, weed) and 6-class (i.e., multiple weed species).
For each task, we provide baselines using two different established architectures: DeepLabv3+ \cite{deeplabv3plus}, and MaskFormer \cite{cheng2021maskformer}. To study the influence of additional imaging bands, we test 3-channel (RGB) and 5-channel (MSI) inputs. We report intersection-over-union (IoU) scores for each class, as well as mean IoU (mIoU) amongst all classes. 

For both architectures, we start with a pre-trained ResNet50 \cite{7780459} backbone, \fblue{only changing the first convolutional layer to 5 input channels with random initialization for the MSI input.}
The models are trained with Adam optimizer \cite{KingBa15}, batch size of 8, and standard data augmentation (i.e. rotation, flipping, random jitter) is applied.
The learning rate is set to 0.001 and 0.00003 for DeepLabv3+ and MaskFormer respectively, and training is done on an Nvidia A100 GPU and an Intel Xeon Ice Lake CPU with 32GB memory, where the final parameters are chosen based on the validation set.

\noindent\textbf{Results.}
The results for the 3-class and 6-class variants can be seen from \cref{tab:semseg_3_class} and \cref{tab:semseg_6_class} respectively.
The results show that both architectures achieve similar mIoU, with DeepLabv3+ having slightly higher scores. \fblue{While the additional NIR and RE bands consistently improve performance, the effects are more pronounced for the more challenging 6-class case. An important observation is that the improvements are most significant for \textit{quickweed} and \textit{weed other}, which are the underrepresented classes. Hence, the results show that MSI can provide additional information where data is scarce. The class-wise improvements also differ between the two chosen architectures. The MSI boost is the highest for \textit{amaranth} (the class with most samples) for MaskFormer, while it's more prominent for \textit{barnyard grass and quickweed} for DeepLabv3+.}

\begin{table}
  \centering
  {\small{
  \begin{tabular}{@{}lcccccc@{}}
    \toprule
    Method & Input & IoU$_\textnormal{bg}$ & IoU$_\textnormal{crop}$ & IoU$_\textnormal{weed}$ & mIoU\\
    \midrule
     DeepLabv3+ \cite{deeplabv3plus} & RGB & 97.97 & 67.93 & 72.08 & 79.33 \\
     & MSI & \bnum{98.45} & \bnum{72.93} & \bnum{77.31} & \bnum{82.90} \\
     MaskFormer \cite{cheng2021maskformer} & RGB & 97.73 & 70.18 & 69.85 & 79.26 \\
      & MSI & 97.99 & 69.49 & 73.33 & 80.27 \\
    \bottomrule
  \end{tabular}
  }}
  \caption{Semantic segmentation scores (\%) for the 3-class setting (uni-weed) on our test set.}
  \label{tab:semseg_3_class}
\end{table}

\begin{table}[t!]
  \centering
  \adjustbox{max width=\columnwidth}{

  \begin{tabular}{@{}lcccccccc@{}}
    \toprule
    Method & Input & IoU$_\textnormal{bg}$ & IoU$_\textnormal{crop}$ & IoU$_\textnormal{am}$ & IoU$_\textnormal{gr}$ & IoU$_\textnormal{qw}$ & IoU$_\textnormal{wo}$ & mIoU\\
    \midrule
    DeepLabv3+ \cite{deeplabv3plus} & RGB & 97.94 & 71.46 & 74.32 & 45.95 & 6.26 & 8.92 & 50.81\\
         & MSI & \bnum{98.37} & \bnum{73.03} & \bnum{76.17} & \bnum{53.55} & \bnum{21.11} & \bnum{10.86} & \bnum{55.52} \\
    MaskFormer~\cite{cheng2021maskformer} & RGB & 97.46 & 68.28 & 68.15 & 41.66 & 4.92 & 3.28 & 47.29 \\
         & MSI & 97.90 & 68.04 & 73.11 & 45.34 & 8.66 & 10.86 & 50.65 \\
    \bottomrule
  \end{tabular}}
  \caption{Semantic segmentation scores (\%) for the 6-class setting on our test set. Crop, am, gr, qw and wo correspond to \textit{maize}, \textit{amaranth}, \textit{barnyard grass}, \textit{quickweed} and \textit{weed other} respectively.}
  \label{tab:semseg_6_class}
\end{table}

\begin{figure}
    \centering
    {\includegraphics[trim={2cm 3cm 6.5cm 3.5cm},clip,width=1\linewidth]{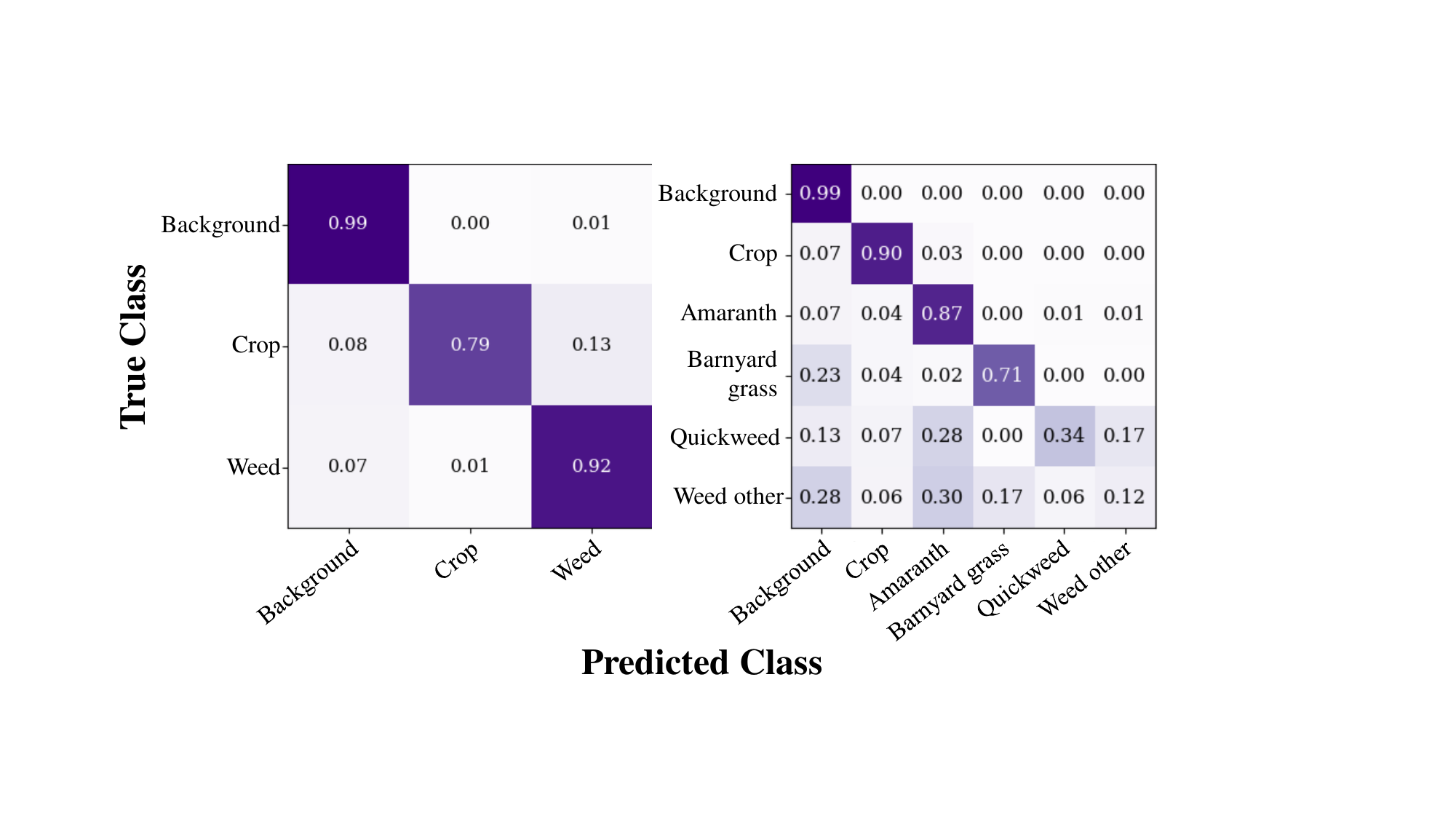}}
    \caption {\textbf{Normalized confusion matrices for semantic segmentation.} Scores reported for  MSI input and DeepLabv3+ model, and uni-weed (left) and multi-weed class cases (right).}
    \label{fig:confusion_matrix}
\end{figure}

Confusion matrices for both cases are shown in \cref{fig:confusion_matrix}
The results for 3-class setting show that 8\% of crops are missed (i.e.~predicted as background) while 13\% are misclassified as weeds.
However, for weeds, almost the entire misclassification comes from false negatives (i.e., weed classified as background), which is especially the case for very small (early growth-stage) weeds as it can be seen from qualitative results in \cref{fig:uncertainty}.
In the 6-class case, the lowest prediction performance is for \textit{quickweed} and \textit{weed other}.
The latter is expected,
as several weed species are pooled together in one class, and they are sometimes classified as one of the other known three categories.
\textit{Quickweed} is classified as \textit{amaranth} by 28\%, which is a very high ratio.
This could be explained by the phenotypes (visual appearances) of the two plants, which are both broad-leaf species with very similar appearances.

\subsubsection{Instance Segmentation}
\label{subsubsec:instance}
\noindent\textbf{Implementation Details and Evaluation Metrics.}
We use MaskFormer to provide baseline results for instance segmentation for a 3-class case and both input modalities. We train the model with the same configuration as in \cref{subsubsec:semantic} and report mAP (mean average precision) scores, which is calculated for IoU thresholds in the range [0.5, 0.95] with a step size of 0.05. Moreover, we report mAP$_\text{50}$ and mAP$_\text{75}$  at 50\% and 75\% IoU respectively. 

\noindent\textbf{Results.} The quantitative scores are reported in \cref{tab:inst_seg}. The two input modalities yield similar results, with MSI being slightly better. The baseline scores for instance-segmentation are overall lower than for other weed segmentation datasets  \cite{weyler2023phenobench}, which can be explained by the relatively large number of plant instances in our dataset (on average more than 78 plants per image), which makes the task challenging, especially for small weeds.

\begin{table}
  \centering
  {\small{
  \begin{tabular}{@{}lcccc@{}}
    \toprule
    Input & mAP & mAP$_\text{50}$ & mAP$_\text{75}$ \\
    \midrule
     RGB & 32.17 & 33.66 & 32.01 \\
     MSI & \bnum{32.53} & \bnum{34.70} & \bnum{33.34} \\
    \bottomrule
  \end{tabular}
  }}
  \caption{Instance segmentation scores (\%).}
  \label{tab:inst_seg}
\end{table}

\subsection{Uncertainty Quantification}
\label{subsec:exp:uq}
\noindent\textbf{Variational Inference with MC Dropout.}
To model uncertainty, we employ VI with MC dropout. It is a widely used UQ method, and has been shown to effectively quantify prediction uncertainty for semantic segmentation of crops and weeds \cite{Celikkan_2023_ICCV}. We use predictive entropy as the uncertainty metric.

\noindent\textbf{Implementation Details and Evaluation Metrics.}
We implement a probabilistic version of DeepLabv3+, by
adding dropout layers with $p_d=$0.5 between each of the four convolutional blocks of the ResNet50 backbone.
During inference, $p_d$ is kept at 0.5, and number of forward passes $K$=5.

We report IoU scores to compare performance with its deterministic counterpart. 
Moreover, we report the expected calibration error (ECE) \cite{pmlr-v70-guo17a}, which is defined as:
\begin{equation*}
    \label{eq:ece}
    \text{ECE} = \sum_{m=1}^M \frac{|B_m|}{n} \left| \text{acc}(B_m) - \text{conf}(B_m) \right|, 
\end{equation*}
where acc and conf represent accuracy and confidence respectively, $n$ is the total number of MC samples and $B_m$ contains predictions that fall into the $m$th bin.
For the deterministic case, the softmax scores are taken as confidence, whereas for the MC dropout version, we take the average over the $K$ forward passes during test time.

\noindent\textbf{Results.} \cref{tab:uq} shows segmentation scores for the 3-class case, for both 3-channel (RGB) and 5-channel (MSI) inputs.
For both input modalities, the probabilistic model is significantly better calibrated and has consistently higher IoU scores.
The improvement in calibration is the most pronounced for the MSI input, where ECE drops to less than half.
Qualitative results including the predictions, as well as error and predictive uncertainty are shown in \cref{fig:uncertainty}. 

Uncertainty, quantified by predictive entropy, is present at the object borders, a common case in many domains due to annotation artifacts and expected noise (captured by aleatoric uncertainty) at the edges \cite{Goan_2023_CVPR, Mukhoti_2023_CVPR, Mukhoti18cvprw}. Moreover, noisy and underrepresented areas, such as the small lumps of soil or stones in the bottom example, have high uncertainty, which can be attributed to both epistemic (not many examples in training data) and aleatoric uncertainty (noisy shadows and varying reflectance values due to changing weather conditions).
Most importantly, the results show a high correlation between the error and high uncertainty regions.
This implies that the model is the most uncertain about pixels that it misclassified, which is a must-have to safely deploy algorithms in real-world applications.

\begin{table}
  \centering
  {\adjustbox{max width=0.8\columnwidth}{
  \begin{tabular}{@{}llcccc@{}}
    \toprule
    Input & Method & IoU$_\textnormal{crop}$ & IoU$_\textnormal{weed}$ & mIoU & $\downarrow$ ECE \\
    \midrule
    RGB & DLv3+ \cite{deeplabv3plus} & 67.93 & 72.08 & 79.33 & 0.0058 \\
     & Prob. DLv3+ & 69.29 & 72.19 & 79.81 & 0.0041 \\
    MSI & DLv3+ \cite{deeplabv3plus} & 72.93 & \bnum{77.31} & 82.90 & 0.0048 \\
     & Prob. DLv3+ & \bnum{73.90} & 76.51 & \bnum{82.94} & \bnum{0.0018} \\
    \bottomrule
  \end{tabular}
  }}
  \caption{\textbf{Comparison of deterministic and probabilistic variants.} Scores for 3-class semantic segmentation on our test set.}
  \label{tab:uq}
\end{table}

\global\long\def\figWidth{0.18\linewidth}
\begin{figure}[ht!]
	\centering
    {\scriptsize
    \setlength{\tabcolsep}{1pt}
	\begin{tabular}{
	>{\centering\arraybackslash}m{0.1cm} 
	>{\centering\arraybackslash}m{\figWidth} 
	>{\centering\arraybackslash}m{\figWidth} 
	>{\centering\arraybackslash}m{\figWidth} 
	>{\centering\arraybackslash}m{\figWidth}
 	>{\centering\arraybackslash}m{\figWidth}
}

	&\includegraphics[clip,trim={{0cm 0cm 0cm 0cm}},width=\linewidth]{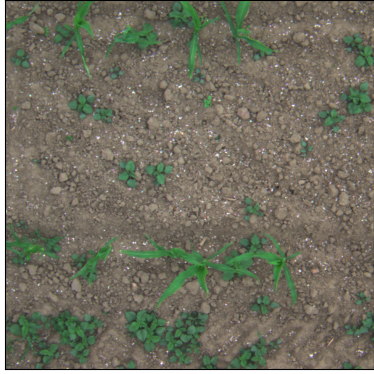}
        &\includegraphics[clip,trim={{0cm 0cm 0cm 0cm}},width=\linewidth]{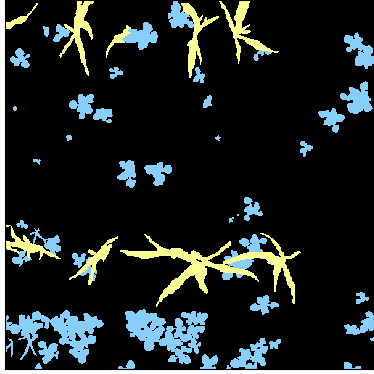}
        &{\includegraphics[clip,trim={{0cm 0cm 0cm 0cm}},width=\linewidth]{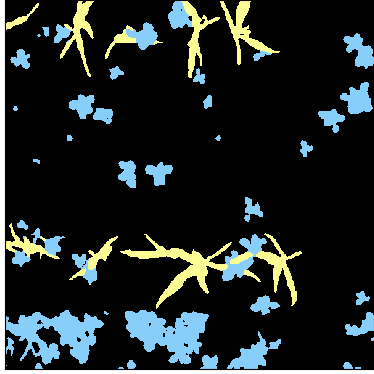}}
         &{\includegraphics[clip,trim={{0cm 0cm 0cm 0cm}},width=\linewidth]{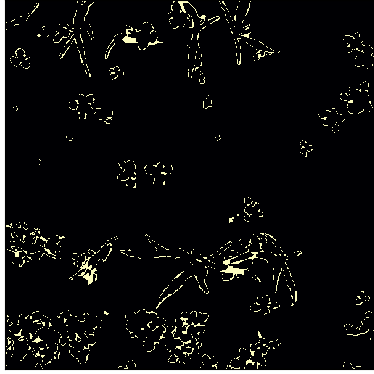}}
        &{\includegraphics[clip,trim={{0cm 0cm 0cm 0cm}},width=1.2\linewidth]{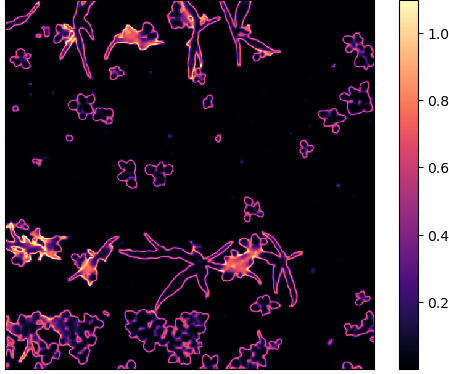}}
        \\

	&\includegraphics[clip,trim={{0cm 0cm 0cm 0cm}},width=\linewidth]{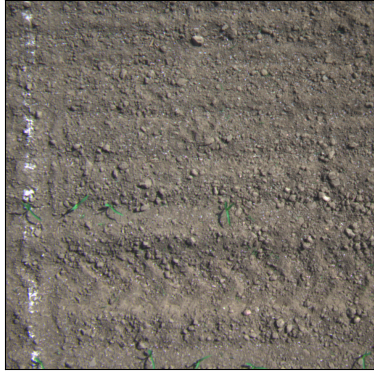}
        &\includegraphics[clip,trim={{0cm 0cm 0cm 0cm}},width=\linewidth]{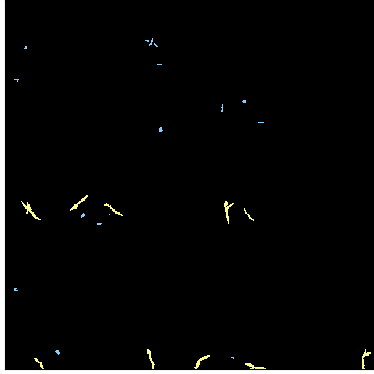}
        &{\includegraphics[clip,trim={{0cm 0cm 0cm 0cm}},width=\linewidth]{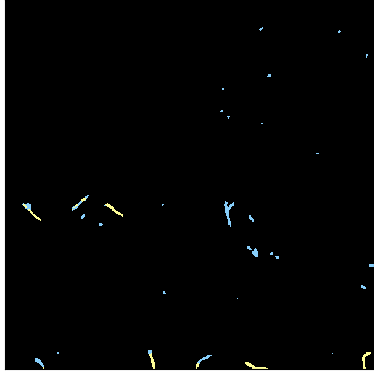}}
         &{\includegraphics[clip,trim={{0cm 0cm 0cm 0cm}},width=\linewidth]{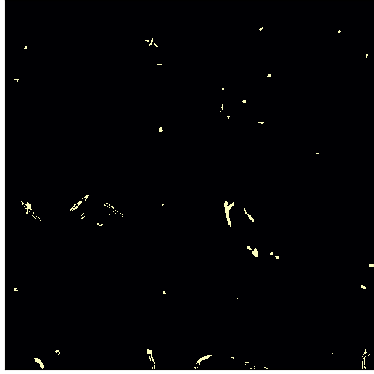}}
        &{\includegraphics[clip,trim={{0cm 0cm 0cm 0cm}},width=1.2\linewidth]{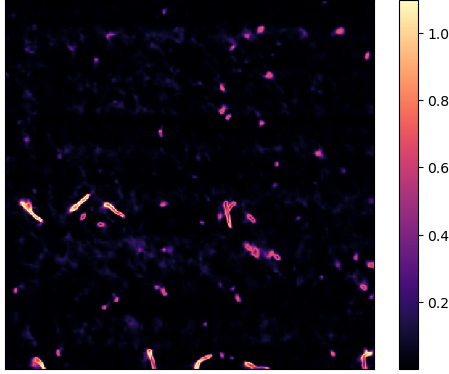}}
        \\

        & \raisebox{0pt}{RGB}
        & \raisebox{0pt}{Ground Truth}
        & \raisebox{0pt}{Predictions}
        & \raisebox{0pt}{Error}
        & \raisebox{0pt}{Entropy}
	\end{tabular}
	}
  \caption{\textbf{Qualitative results for probabilistic crop and weed segmentation for MSI input.} The uncertainty is high in regions that are misclassified, which is a desired and useful information.}
  \label{fig:uncertainty}
\end{figure}

\definecolor{crop}{HTML}{caca78}

\begin{figure*}
    \centering
    {\includegraphics[trim={0cm 9cm 0cm 0cm},clip,width=15cm]{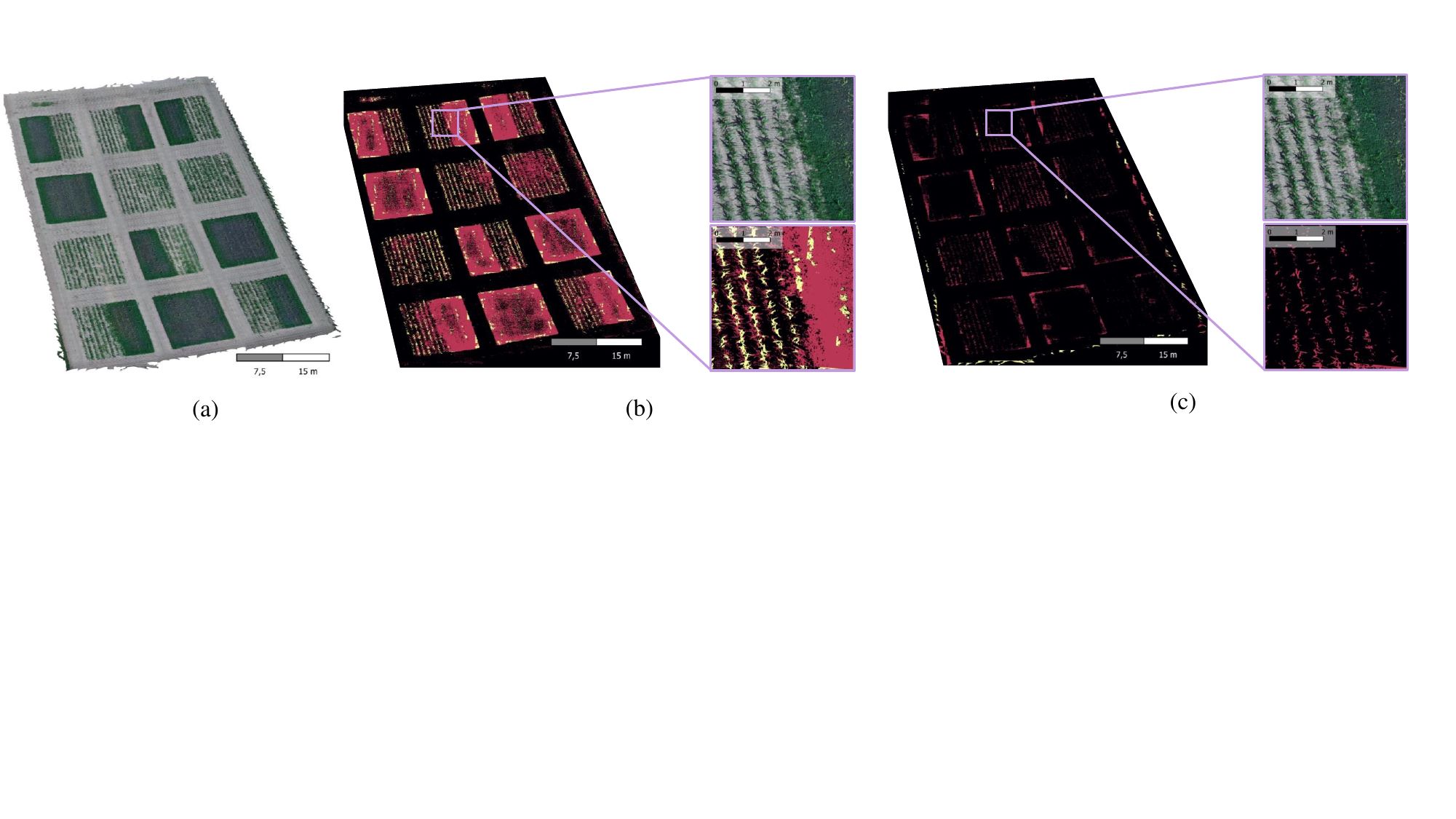}}

  \caption{\textbf{Weed cover monitoring on OOD data.} (a) Full orthomosaic of Maize2024 field, where control areas (i.e.~not sprayed) have full plant cover because the weeds took over. Semantic segmentation results (\wpink{weed} vs. \textcolor{crop}{crop}) for when (b) trained on WeedsGalore and (c) trained on MaizeOrWeed. A sample patch with its predictions is shown zoomed in. Best viewed on colored screen and zoomed in.}
  \label{fig:ood}
\end{figure*}

\subsection{Application to Unseen Data}
\label{subsec:generalization}

This section quantitatively demonstrates the need for a reference dataset, designed specifically for weed monitoring in maize fields using UAVs, and showcases the generalization potentials of WeedsGalore to unseen data. To this end, we introduce an additional test field named \textit{Maize2024}. This OOD data includes recordings from a different field, year, plant cover, and unseen (i.e.~later) growth stages. These are the domain shifts that we expect in our application. Maize2024, an agricultural area in Marquardt, Potsdam, Germany, is an experimental field where half of the field (12 out of 24 patches) are assigned as \textit{control} areas (ie. no herbicide treatment) whereas the other half are \textit{test} (ie. sprayed with herbicides). The data is collected with the same drone after the treatment had its effect on the field, which means that weeds were mostly eliminated in the sprayed (\textit{test}) patches whereas \textit{control} areas had full plant cover (i.e. complete weed infestation). 

The experimental setup enables informative qualitative evaluation. A well-performing model is expected to predict almost full weed cover in \textit{control} patches, and little in the sprayed \textit{test} areas. For the task of 3-class semantic segmentation, we train DeepLabv3+ separately on the following datasets: WeedsGalore (ours), PhenoBench \cite{weyler2023phenobench} (same acquisition mode, different crop), CropAndWeed \cite{Steininger_2023_WACV} (includes maize along others, different acquisition mode), and another variant of CropAndWeed, which we refer to as \textit{MaizeOrWeed}, a subset of CropAndWeed (1753 images), including only scenes with maize. \cref{fig:ood} shows qualitative predictions on the large-scale orthomosaic of Maize2024 for models trained on WeedsGalore and MaizeOrWeed. The former clearly captures the high weed cover in \textit{control} areas, while the latter completely fails to segment any vegetation.

To get quantitative scores, from the  Maize2024 orthomosaic \cref{fig:ood}(a) we annotate 6 cropped patches corresponding to 90 m$^2$ ground area, each of size $1000 \times 1100 px$. \cref{tab:maize24} shows the scores for 3-class semantic segmentation. The model trained on WeedsGalore (our dataset) achieves 52.55\% mIoU on OOD Maize2024 data, 23.22\% higher than the runner-up, confirming the qualitative results. The scores confirm that the acquisition mode or crop type alone isn't sufficient to handle domain shifts, and datasets targeting UAV-based weed-maize monitoring are needed.

\FloatBarrier
\begin{table}
  \centering
  \adjustbox{max width=0.95\columnwidth}{
  \arrayrulecolor{black}
  \color{black}
  \begin{tabular}{@{}l|lcc{|}ccccc@{}}
    \toprule
    Model & Source & UAV & Maize & IoU$_\textnormal{bg}$ & IoU$_\textnormal{crop}$ & IoU$_\textnormal{weed}$ & $\uparrow$ mIoU & $\downarrow$ ECE\\
    \midrule
    DLv3+ & PhenoBench  & \ding{51} & \ding{55} & 56.94 & 30.80 & 0.25 & 29.33 & 0.4519 \\
    DLv3+ & CropAndWeed & \ding{55} & \ding{51} & 51.59 & 0.00  &  7.02 & 19.54 & 0.4130 \\
    DLv3+ & MaizeOrWeed & \ding{55} & \ding{51} & 50.25 & 0.00 & 0.21 & 16.82 & 0.5154 \\
     \midrule
    DLv3+ & WeedsGalore (Ours) & \ding{51} & \ding{51}  & 67.10 & 35.70 & 54.84  & 52.55 & 0.1290  \\
    Prob. DLv3+ & WeedsGalore (Ours) & \ding{51} & \ding{51} & \bnum{68.65} & \bnum{50.37} & \bnum{57.32} & \bnum{58.78} & \bnum{0.0887}  \\
    \bottomrule
  \end{tabular}
  }
  \caption{\fblue{\textbf{Results on Maize2024 data.}
  IoU for 3-class semantic segmentation of DeepLabv3+ trained on different datasets, including the probabilistic variant for WeedsGalore.}}
  \label{tab:maize24}
\end{table}

\global\long\def\figWidth{0.17\linewidth}

\definecolor{crop}{HTML}{caca78}
\definecolor{weed1}{HTML}{87CEFA}
\definecolor{weed2}{HTML}{B0E0E6}
\definecolor{weed3}{HTML}{ccc0c4}
\definecolor{weed_other}{HTML}{CCCCFF}

\begin{figure}[ht!]
	\centering
    {\scriptsize
    \setlength{\tabcolsep}{2pt}
	\begin{tabular}{
	>{\centering\arraybackslash}m{0.2cm} 
	>{\centering\arraybackslash}m{\figWidth} 
	>{\centering\arraybackslash}m{\figWidth} 
	>{\centering\arraybackslash}m{\figWidth} 
        >{\centering\arraybackslash}m{\figWidth}
        >{\centering\arraybackslash}m{\figWidth} 
        >{\centering\arraybackslash}m{\figWidth}
        }

        \rotatebox{90}{\makecell{\textit{control/test}}}
	&\includegraphics[clip,trim={{0cm 0cm 0cm 0cm}},width=\linewidth] 
          {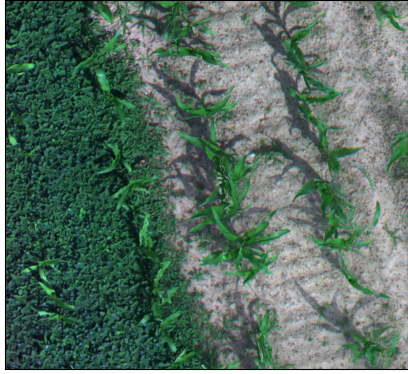}
          
        &\includegraphics[clip,trim={{0cm 0cm 0cm 0cm}},width=\linewidth] 
         {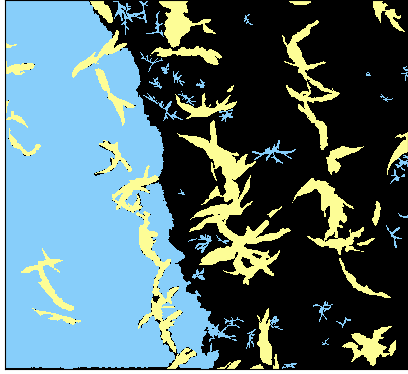}

        &\includegraphics[clip,trim={{0cm 0cm 0cm 0cm}},width=\linewidth] 
         {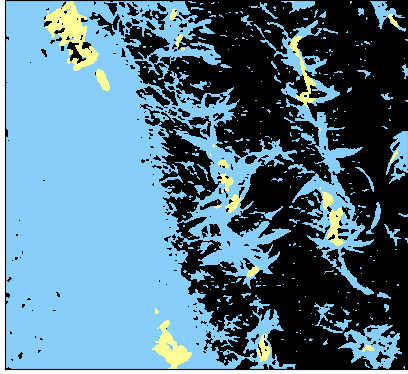}
         
        &\includegraphics[clip,trim={{0cm 0cm 0cm 0cm}},width=\linewidth] 
         {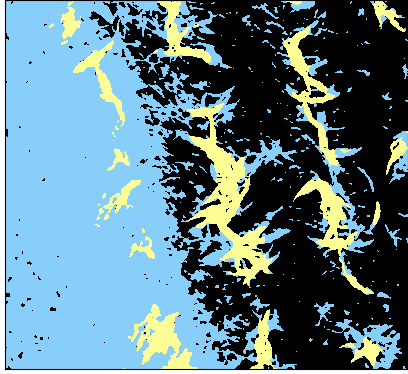}
         
        &\raisebox{0pt}{\includegraphics[clip,trim={{0cm 0cm 0cm 0cm}},width=1.2\linewidth]{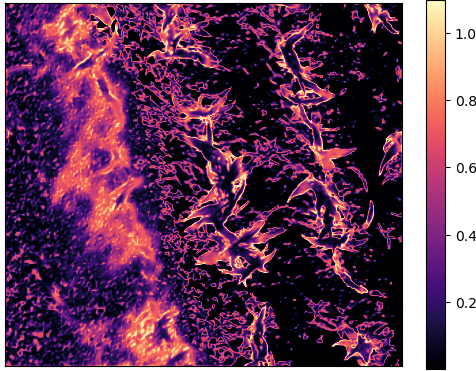}}

        \\

        & \raisebox{0pt}{Input}
        & \raisebox{0pt}{Ground truth}
        & \raisebox{0pt}{\parbox{1.5cm}{Prediction \\ \centering  (det., 51\%)}}
        & \raisebox{0pt}{\parbox{1.5cm}{Prediction \\ \centering (prob., 67\%)}}
        & \raisebox{-4pt}{Entropy}

	\end{tabular}
	}
  \caption{\fblue{\textbf{Predictions on Maize2024.} Left side of the image from \textit{control}, and right from \textit{test} area. Predictions from both deterministic and probabilistic models are captioned with mIoU scores. The uncertainty map (i.e.~entropy) produced by the probabilistic model highlights the challenging areas.}}
  \label{fig:uq_maize_2024}
\end{figure}

We deploy the probabilistic variant of the model on Maize2024 to show the significance and potentials of probabilistic approaches on unseen data. As it can be seen from \cref{tab:maize24}, the probabilistic variant yields a 15\% boost in crop IoU. This increased segmentation performance is accompanied by improved model calibration and informative uncertainty scores. \cref{fig:uq_maize_2024}, which includes sample segments from both \textit{control} and \textit{test} areas, qualitatively demonstrates that the deterministic model tends to misclassify soil and plant shadows, and parts of maize leaves as weeds, while probabilistic variant handles those challenging areas better. A common failure case is when there is full weed coverage (i.e.~\textit{control} areas), where many crops are missed by the model. In those areas, the uncertainty is very high, which could be useful in a real-world scenario (e.g.~even though the area is segmented as \textit{weed}, the weeding system may decide \textit{not spray} to not to risk potential crops as the uncertainty is very high).

\section{Conclusion}
\label{sec:conclusion}
We introduce a novel UAV dataset for crop and weed segmentation in maize fields.
The dataset includes multitemporal images from different growth stages and five spectral bands, along with high-quality instance and semantic annotations.
This reference data can inspire further research in aerial weed identification and monitoring.
We present baseline results for semantic and instance segmentation tasks, including probabilistic methods to quantify uncertainty, improve model calibration, and demonstrate the approach's applicability for weed cover monitoring on OOD data. We believe that the increased availability of more datasets targeting different species and using various sensing technologies will enhance model development, aiding automated weeding and monitoring systems.

\section{Acknowledgements}
\label{sec:acknowledgements}
The authors acknowledge the support of the Helmholtz Einstein International Berlin Research School in Data Science (HEIBRiDS) and the GFZ Potsdam. This work utilized high-performance computing resources made possible by funding from the Ministry of Science, Research and Culture of the State of Brandenburg (MWFK) and are operated by the IT Services and Operations unit of the Helmholtz Centre Potsdam.

{\small
\bibliographystyle{ieee_fullname}
\bibliography{bib}
}

\end{document}